\documentclass[sigconf]{acmart}

\usepackage{booktabs} 

\setcopyright{none}

\usepackage{import}
\usepackage{amsmath}
\usepackage{graphicx}
\usepackage{multirow}

\usepackage{hyperref}

\begin{document}
\title{Multi-label Classification of User Reactions in Online News}

\author{Zacarias Curi}
\affiliation{%
  \institution{Pontif\'icia Universidade Cat\'olica do Paran\'a (PUCPR)}
  \city{Curitiba, PR} 
  \state{Brazil}
}
\email{zacarias@ppgia.pucpr.br}

\author{Alceu de Souza Britto Jr}
\affiliation{%
  \institution{Pontif\'icia Universidade Cat\'olica do Paran\'a (PUCPR)}
  \city{Curitiba, PR} 
  \state{Brazil}
}
\affiliation{%
  \institution{Universidade Estadual de Ponta Grossa (UEPG)}
  \city{Ponta Grossa, PR} 
  \state{Brazil}
}
\email{alceu@ppgia.pucpr.br}

\author{Emerson Cabrera Paraiso}
\affiliation{%
  \institution{Pontif\'icia Universidade Cat\'olica do Paran\'a (PUCPR)}
  \city{Curitiba, PR} 
  \state{Brazil}
}
\email{paraiso@ppgia.pucpr.br}


\renewcommand{\shortauthors}{Z. Curi, A.S. Britto Jr, E.C. Paraiso}

\begin{abstract}
The increase in the number of Internet users and the strong interaction brought by Web 2.0 made the Opinion Mining an important task in the area of natural language processing. Although several methods are capable of performing this task, few use multi-label classification, where there is a group of true labels for each example. This type of classification is useful for situations where the opinions are analyzed from the perspective of the reader, this happens because each person can have different interpretations and opinions on the same subject. This paper discuss the efficiency of problem transformation methods combined with different classification algorithms for the task of multi-label classification of reactions in news texts. To do that, extensive tests were carried out on two news corpora written in Brazilian Portuguese annotated with reactions. A new corpus called BFRC-PT is presented. In the tests performed, the highest number of correct predictions was obtained with the Classifier Chains method combined with the Random Forest algorithm. When considering the class distribution, the best results were obtained with the Binary Relevance method combined with the LSTM and Random Forest algorithms.
\end{abstract}

%


\keywords{multi-label classification, opinion mining, LSTM}

\maketitle

\section{Introduction}
\label{sec:introduction}

The success of Web 2.0 provides a constant generation of a large amount of textual data. The sites are very interactive. This characteristic combined with the cultural diversity of users ensures that different organizations are interested in the information contained in those texts. In this scenario, the task of opinion mining became popular in the area of Natural Language Processing (NLP) since it can provide the tools for information extraction and knowledge acquisition.  Among the existing techniques, Deep Learning has been achieving good results for the classification task in cases where there is a large amount of data. An example of this is the application of the algorithm Long Short-Term Memory (LSTM) in the analysis of texts generated by the Web 2.0 published in \cite{day2017deep} and \cite{wang2016attention}.


Most of the opinion mining research perform the single-label classification, in which only one label is considered for each text. This type of classification is efficient in cases where the purpose is to analyze the opinion expressed by the writer. However, there are situations where the goal is to analyze the text from the reader's perspective, that is how the reader reacted when was reading the text. The term reaction is defined in this work as the attitude or sensation acquired by a person upon receiving a stimulus from an external source. The reaction can be presented as a component of the emotions. Desmet \cite{desmet2003measuring} defines that emotions can be treated as a multifaceted phenomenon, consisting of behavioural reactions, expressive reactions, physiological reactions and subjective feelings. In this work, we analyze a corpus annotated with expressive reactions and with emotions. As the corpora used are annotated from the perspective of the reader, it is necessary to use the multi-label classification, in which several reactions are considered simultaneously for the same text. This type of classification is necessary because each person has their individuality, which generates different reactions and consequently different emotions.


The task of multi-label classification can be accomplished through an adaptation of the classification algorithm or a transformation in the problem. The algorithm adaptation methods consider performing transformations in the traditional single-label classification methods to allow the use of multi-label problems. Problem transformation methods consider to transform a given problem into one or more single label problems \cite{zhang2014review}.


This work aims to compare some problem transformation methods combined with different induction algorithms for the task of classifying reactions in texts. This work also aims to perform the task of multi-label classification with the use of several binary LSTM classifiers. In this way, we try to verify if the use of the LSTM allows better results than the traditional methods of classification, as well as those obtained in several works with classification of single label.


The development of efficient strategies to classify reactions in texts is necessary to guarantee good results in the applications of this task. The identification of reactions in news can be used to recommend new news to users, it's also can be applied to select appropriate ads or offers to users. This task is also useful for making decisions about the subject covered in a news story and about the news with the greatest potential for the newspaper.


This paper presents three main contributions. We compared the efficiency of some traditional methods of problem transformation considering different induction algorithms such as Support Vector Machine (SVM), Naive Bayes (NB), and Random Forest (RF). To the best of our knowledge, this is the first work to apply the LSTM algorithm with a problem transformation method for the task of classifying reactions in texts. Another contribution of this work is the introduction of a new corpus of online news written in Portuguese, labeled with user reactions. 


The remainder of this paper is organized as follows. Section 2 presents some related work. Section 3 describes the methods and algorithms evaluated. The experiments and corresponding results are presented in Section 4 and 5, respectively. Finally, in Section 6 we present our conclusions and future work.

\section{Related Work}
\label{sec:relatedworks}

Most of the opinion mining works perform the single label classification. However, Liu and Chen \cite{liu2015multi} present the analysis of texts extracted from a Chinese microblog annotated in multi-label form. The authors presented a comparison with 11 methods of problem transformation and algorithm adaptation to classify these texts. Another way to accomplish this task is presented by Song and colleagues \cite{song2015build}, in which lexicons were used.

    
Another work using lexicons is presented by Phan, Shindo and Matsumoto \cite{phan2016multiple}. The authors report the creation of a new resource using a Recurrent Neural Network. The feature created is used for multi-label classification of Plutchik's basic emotions in transcripts of film dialogues. An approach using Deep Learning with a problem transformation technique is presented by Wang, Ren and Miao \cite{wang2016multi}. They proposed a method based on Convolutional Neural Network (CNN) for the multi-label classification of emotions in sentences of microblogs in Chinese.

    
Most opinion mining works use copora annotated from the writer's perspective. In the paper presented by Pool and Nissim \cite{pool2016distant}, the authors use a corpus annotated from the perspective of the reader, using Facebook\footnote{\url{https://www.facebook.com/}} messaging reactions. Although this work performs a single label classification, the used corpus  contains more than one emotion associated with each text. Bhowmick and colleagues \cite{bhowmick2009multi} used an algorithm adaptation technique called ML$k$NN to classify four emotions into a news corpus labeled from the perspective of the reader. Zhang et al. \cite{zhang2015multi} present a new framework for classifying a corpus from the same perspective.

    
As for the single-label classification, most of the works existing in the literature perform the classification of texts in English or Chinese. Zwaan and colleagues \cite{van2015heem} present the use of the Problem Transformation methods BR and RA$k$EL with the SVM algorithm for the classification of texts in Dutch. Another explored language is Japanese. Duan and colleagues \cite{duan2014separate} report the use of crowdsourcing for annotating two children's stories in that language. The authors also present two techniques based on the Problem Transformation methods called BR and LP with the NB classification algorithm. In this work, we use a corpus in Brazilian Portuguese annotated from the perspective of the reader. The novelty is related to the use of Deep Learning with a Problem Transformation Method for classification. The LSTM and the Problem Transformation methods used in this work are presented in the next section.


\section{Classification Methods}
\label{sec:Methods}
This section briefly presents the multi-label classification methods used in the experimental part of this work. It also present the LSTM algorithm. 

\subsection{Multi-label classification}
The most common approaches to traditional supervised learning tasks perform single-label classification. In this type of classification, each sample is represented by only one label. Considering $\lambda$ as a single label for an instance of the database used and $\mathit {L}$ as the class set of the problem, we have the classification called binary for cases where $ |\mathit{L}| = 2$. In cases where $ |\mathit {L} | > 2$ the classification is called multi-class. Different from binary and multi-class classification, where there is only one label $\lambda$ for each instance, the multi-label classification accepts a set of labels $\mathit{Y}$ to represent each instance, such that $\mathit{Y} \subseteq \mathit{L}$ \cite{tsoumakas2007multi}. In short, multi-label problems can be defined as situations where there is a set of true labels for each instance of the problem, and for at least one instance the set has more than one label. Currently, two groups of methods to solve this type of problem can be found in the literature: Problem Transformation and Algorithm Adaptation \cite{zhang2014review}.


The Problem Transformation techniques consist of transforming the multi-label classification problem into one or more single-label sorting or classification problems. One way to accomplish this task is to create an independent binary classifier for each label of the problem, using the method called Binary Relevance (BR). The main problem of the BR method is that it does not consider the dependency between the labels, thus ignoring some characteristics of the problem \cite{zhang2017binary}. One way to solve this is through the Classifier Chains (CC) method. The CC uses the output of a binary classifier as an input attribute to the next, thereby creating a link between binary classifiers and adding the relationship between classes in problem resolution \cite{read2011classifier}. The main problem of this method is the choice of the best order of the classifiers.


An alternative to the transformation of the multi-label problem into several binary problems, as performed in the BR and CC methods, is the transformation into a multiclass problem. An example of this is the Label Powerset (LP) method. This method creates a new label for each label combination in the training database. The main advantage of the strategy used by the LP method is the need for only one classifier. By contrast, the LP method can generate many new classes depending on the characteristics of the database used \cite{tsoumakas2007random}.


One way to reduce the problem of generating new classes in the LP method is to create class groups through a class ensemble method. One of these methods is the Random $\textit{k}$-Labelsets (RA$k$EL), which creates a class ensemble for the LP method. The RA$k$EL method divides the initial set of labels into $m$ random subsets with $k$ classes called label sets. After this division, the label Powerset method is used to perform the transformation of the problem and enable the training \cite{tsoumakas2007random}. An ensemble is also performed by the Hierarchy Of Multilabel classifiers (HOMER), where the multi-label problem is transformed into a hierarchical problem. The main advantage of the hierarchical division created is the use of fewer classes in each classifier and the more balanced distribution between these classes \cite{tsoumakas2008effective}.


Besides the division of the multi-label problem into one or more classification problems, it is possible to carry out the transformation of the multi-label problem into a ranking problem. One of the methods that use this strategy is the Calibrated Label Ranking (CLR), introduced by \cite{furnkranz2008multilabel}. The basic idea of this method is to transform the multi-label problem into a label ranking problem, where the position of the labels is decided on the basis of peer-comparison techniques and is used to perform the classification. For this, the traditional Label Ranking algorithm is used to perform the ordering of the labels based on their relevance, after the ordering a calibration of labels is added, allowing the separation of the relevant labels from the irrelevant labels.


An alternative to problem transformation methods are the algorithm adaptation methods. These methods are defined as all traditional data mining algorithms that are adapted to work directly with a multi-label problem \cite{zhang2014review}. One of these changes is ML$k$NN, presented by \cite{zhang2005k} and \cite{zhang2007ml}. In these works, the authors perform an adaptation of the KNN algorithm to allow the use of multi-label data. In the first step of the ML$k$NN algorithm, all the k nearest neighbors of each instance are identified. After this identification, statistical information obtained from the neighbor's label sets is used for use with the maximum a posteriori principle, which is used to determine the labels. The ML$k$NN algorithm and the other methods of algorithm adaptation in the literature are directly linked to its origin algorithm. Unlike these methods, problem transformation methods can be used with any single-label algorithm. The following section presents one of these algorithms, the LSTM.


\subsection{Long Short-Term Memory}

\begin{figure}[!htb]
    \centering
    \includegraphics[width=0.3\textwidth]{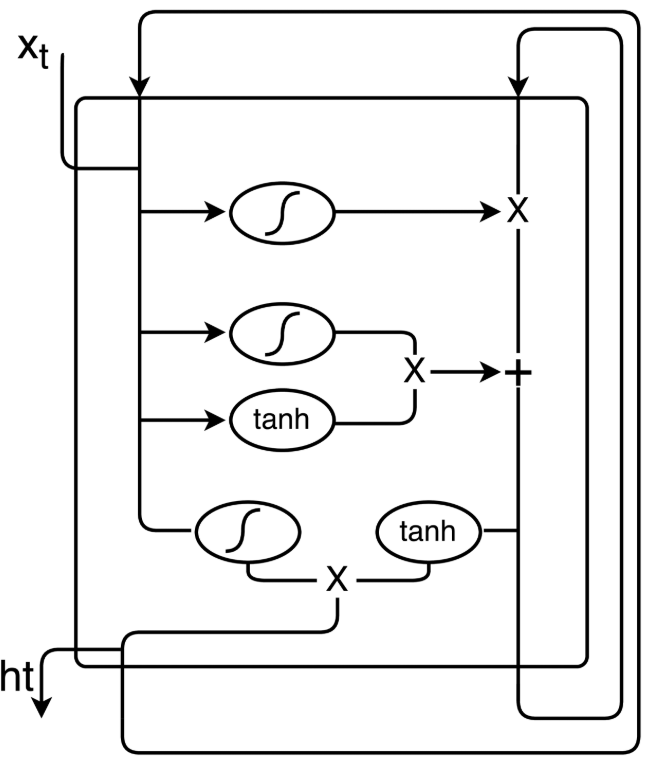}
    \caption{Structure of a Long Short-Term Memory cell.}
    \label{architectureLSTM}
\end{figure}

The Long Short-Term Memory (LSTM) algorithm, initially presented in \cite{hochreiter1997long}, is a type of Recurrent Neural Network (RNN) capable of using long-term stored information for training. In conventional RNNs, it is possible to make a connection with some previous information, but this algorithm is unable to deal with distant information. To solve the problem of the dependence of terms, the model establishes a new structure called memory cell, shown in Figure \ref{architectureLSTM}. This structure is composed of an input gate, neurons with recurrent connections, a forget gate and an output gate. The first step of the LSTM is to use a sigmoid layer to decide what information will be discarded from the current cell. After choosing what will be discarded another sigmoid layer is used to decide what new information should be stored in the cell. Then a $tanh$ selects candidates to be stored in a vector. After the vector creation, a sigmoid function is used to decide which information is best for the next cell. With all the steps operated the old cell is updated and the process is performed again with the new data.


The steps performed by the LSTM can also be represented by the Equations \ref{eqlstm1} to \ref{eqlstm5}. In these equations, the subscript characters represent vectors and the characters in uppercase arrays. In the notation used, $f$ represents the forget gate, $i$ represents the input gate, $o$ the output gate, $c$ represents the memory cell and $h$ the LSTM unit. The matrices are $W$, which stores the input weights and $U$, which stores the recurring connections.


\setlength{\belowdisplayskip}{0pt} \setlength{\belowdisplayshortskip}{0pt}
\setlength{\abovedisplayskip}{0pt} \setlength{\abovedisplayshortskip}{0pt}

\begin{equation}
\label{eqlstm1}
    f_{t} = \sigma_{g}(W_{f}x_{t}+U_{f}h_{t-1}+b_{f})
\end{equation}

\begin{equation}
\label{eqlstm2}
    i_{t} = \sigma_{g}(W_{i}x_{t}+U_{i}h_{t-1}+b_{i})
\end{equation}

\begin{equation}
\label{eqlstm3}    
    o_{t} = \sigma_{g}(W_{o}x_{t}+U_{o}h_{t-1}+b_{o})
\end{equation}

\begin{equation}
\label{eqlstm4}    
    c_{t} = f_{t}\circ c_{t-1}+i_{t}\circ \sigma_{c}(W_{c}x_{t}+U_{c}h_{t-1}+b_{c})
\end{equation}

\begin{equation}
\label{eqlstm5}
    h_{t} = o_{t}\circ \sigma _{h}(c_{t})
\end{equation}

\vspace{4mm}

The next section gives the details concerning the experiments performed in this work.

\section{Experiments}
\label{sec:experiments}
This section presents the experimental protocol followed during experimentation.

\subsection{Portuguese news corpora}

\begin{figure}[!htb]
    \centering
    \includegraphics[width=0.45\textwidth]{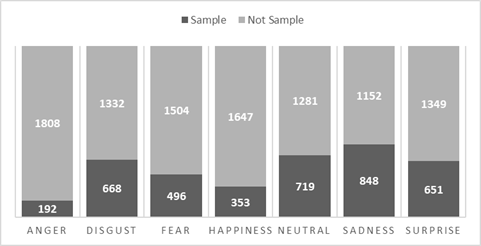}
    \caption{Class Distribution of the G1 corpus}
    \label{dist_g1}
\end{figure}

To evaluate our approach, two news corpora were used. The corpus called G1 was initially presented by Dosciatti and colleagues in \cite{dosciatti2015anotando}. This corpus is composed of 2,000 titles and headlines of news extracted from the website G1\footnote{\url{http://g1.globo.com/}}. The news is annotated with the six basic emotions presented by Ekman \cite{ekman1992argument} and the neutral class for cases where none of the emotions were present in the document. The classes used were: anger, disgust, fear, happiness, sadness, surprise and neutral. Originally, each news was labeled by two annotators, where each annotator identified the primary and secondary emotion of each news. In cases of a tie, a third annotator was consulted to define the primary emotion. The version of the corpus used in this work considers all the emotions selected by the annotators, being able to simultaneously have up to four labels. The number of examples of each label is shown in Figure \ref{dist_g1}. As can be seen in this figure, the G1 corpus is unbalanced, with 192 examples for the minority class (anger) and 848 for the majority class (sadness). The label cardinality of this corpus is 1.964 and the label density is 0.280.


\begin{figure}[!htb]
    \centering
    \includegraphics[width=0.45\textwidth]{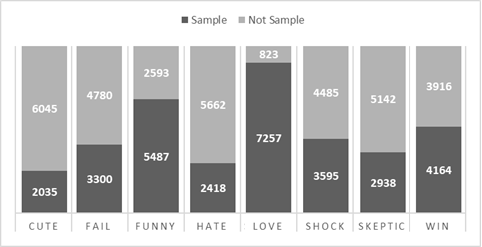}
    \caption{Class distribution of the BFRC-PT corpus}
    \label{dist_bfrc_pt}
\end{figure}

In addition to the G1 corpus, we are presenting a new corpus named BuzzFeed Reactions Corpus (BFRC-PT) consisting in 8,080 entertainment news written in Brazilian Portuguese, collected from the Brazilian version of BuzzFeed\footnote{\url{https://www.buzzfeed.com/?country=pt-br}}. The news was annotated with the vote of the users. During the corpus collection (the first quarter of 2017), the site provided a field for the users to express their reactions for each news read. The eight labels of the presented corpus are defined based on these reactions, being: cute, fail, funny, hate, love, shock, skeptic and win. Because BuzzFeed is focused on entertainment, many news articles feature only pictures or videos, with no textual information. As the focus is the text, these news were discarded. Another change was the application of a threshold to discard the labels with few votes. Analyzing the votes, it was possible to observe inconsistencies, especially in the most popular news. For this reason, all labels with less than 3\% of the total sum of the news labels were deleted. Even with the application of this threshold, the BFRC-PT has some degree of imbalance. Figure \ref{dist_bfrc_pt} presents the distribution of the labels of this corpus. The label cardinality of this corpus is 3.861 with the label density of 0.483. The corpus BFRC-PT can be accessed at link\footnote{\url{https://www.ppgia.pucpr.br/~paraiso/mineracaodeemocoes/recursos.php}}.


\subsection{Methods and Algorithms}
One of the main objectives of this work is to verify the efficiency of the LSTM algorithm when used with the Binary Relevance method in comparison to the traditional approaches in the task of multi-label classification of reactions in texts. The Problem Transformation methods BR, CC, CLR, HOMER, LP and RA$k$EL were used with the NB, RF and SVM algorithms to be compared with the BR method with the LSTM algorithm and the adapted algorithm ML$k$NN. The parameters of the HOMER were defined based on the work \cite{tsoumakas2008effective}. The RA$k$EL has been tested with all available settings. All other methods have been tested with their default configuration. The experiments were performed with the implementations available in the Meka\footnote{\url{http://waikato.github.io/meka/}} and Mulan\footnote{\url{http://mulan.sourceforge.net/}} software and with an implementation of the LSTM algorithm in the TensorFlow\footnote{\url{https://www.tensorflow.org/}} framework. Due to unbalance, the databases were divided with 3-folds cross-validation. This division allows a greater number of examples of the minority classes in the test database, allowing an improvement in the evaluation.


For the pre-processing of the data used with the traditional algorithms, all texts were converted to the lowercase and special characters were removed. All words found in the stopwords list provided by SnollBall\footnote{\url{http://snowballstem.org/}} were removed. A stemming, also provided by the snowball system, was applied to extract the radicals from the words. All links, emails, numbers, currency symbols and percentages were replaced by tokens. Finally, the TF-IDF (term frequency-inverse document frequency) method was applied to represent the words in vector form. For the LSTM algorithm, no changes were made to the words. For this algorithm, we used the embedding method word2vec with the vector pre-trained by Hartmann and Colleagues \cite{hartmann2017portuguese}.


\subsection{Evaluation Metrics}
For the analysis of the used methods, two multi-label evaluation metrics were used: the hamming loss and the micro-F1. The hamming loss metric is defined by the Equation \ref{eqhl}, where $\triangle$ implies the symmetric difference between two sets, $X$ represents the test set, $L$ the problem classes, $h(x_i)$ is the classifier prediction for the instance $x_i$, while $Y_i$ corresponds to its label.

            
\begin{equation}
    HL(h) = \frac{1}{|X|} \frac{1}{|L|} \sum_{i=1}^{|X|} |h(x_{i}) \:\triangle \: Y_{i}|
    \label{eqhl}
\end{equation}

\vspace{4mm}

The metric F1$_{ml}$ constitutes the adaptation of the existing metric for single-label problems to multi-label problems. Like the original metric, the F1$_{ml}$ represents the harmonic mean between precision and recall being efficient to measure cases where the database is unbalanced. The adaptation occurs in the way the predicted values are calculated, when the values of each label are summed in Equation \ref{eqmi}. After the definition of the values of the confusion matrix, the traditional $F^{\beta}$ metric is applied. This metric is presented in the Equation \ref{eqfm2}.


\begin{equation}
    B_{micro}(h) = B \left( \sum_{j=1}^{|L|} VP_j, \sum_{j=1}^{|L|} FP_j, \sum_{j=1}^{|L|} VN_j, \sum_{j=1}^{|L|} FN_j \right) 
    \label{eqmi}
\end{equation}

\vspace{4mm}

\begin{equation}
     F^{\beta}(h) = \frac{ (1+\beta^2) \: . \: VP_j }{ (1+\beta^2) \: . \: VP_j + \beta^2 \: . \: FN_j + FP_j }
    \label{eqfm2}
\end{equation}

\vspace{4mm}

\section{Results and Discussion}
\label{sec:results}

The corpora used have different amounts of examples and texts with different sizes. These differences in the characteristics of the corpora can generate differences in the results of the classification methods. Table \ref{results} presents the results obtained for both corpora.


\begin{table}[!h]
    \renewcommand{\arraystretch}{1.3}
    \caption{Results obtained}
    \label{results}
    \centering
    \resizebox{3.5in}{!}{%
    \begin{tabular}{|c|c|c|c|c|c|}
    \hline
    \multirow{2}{*}{\textbf{Classifier}} & \multirow{2}{*}{\textbf{Method}} & \multicolumn{2}{c|}{\textbf{micro F1}} & \multicolumn{2}{c|}{\textbf{hamming loss}} \\ \cline{3-6} 
     &  & \textbf{G1} & \textbf{BFRC-PT} & \textbf{G1} & \textbf{BFRC-PT} \\ \hline
    \textbf{KNN} & \textbf{ML$k$NN} & 0.46045 & 0.63438 & 0.31458 & 0.35809 \\ \hline
    \textbf{LSTM} & \textbf{BR} & \textbf{0.56071} & 0.64630 & 0.25416 & 0.35981 \\ \hline
    \multirow{6}{*}{\textbf{NB}} & \textbf{BR} & 0.52529 & 0.60962 & 0.26621 & 0.37651 \\ \cline{2-6} 
     & \textbf{CC} & 0.54493 & 0.61479 & 0.27793 & 0.37953 \\ \cline{2-6} 
     & \textbf{CLR} & 0.54070 & 0.63427 & 0.28243 & 0.37361 \\ \cline{2-6} 
     & \textbf{HOMER} & 0.48756 & 0.60762 & 0.28749 & 0.37851 \\ \cline{2-6} 
     & \textbf{LP} & 0.44118 & 0.61481 & 0.31343 & 0.37143 \\ \cline{2-6} 
     & \textbf{RAkEL} & 0.54506 & 0.63103 & 0.27700 & 0.36465 \\ \hline
    \multirow{6}{*}{\textbf{RF}} & \textbf{BR} & 0.56057 & \textbf{0.65713} & 0.25092 & 0.33685 \\ \cline{2-6} 
     & \textbf{CC} & 0.50541 & 0.61007 & \textbf{0.22678} & \textbf{0.33048} \\ \cline{2-6} 
     & \textbf{CLR} & 0.55483 & 0.65506 & 0.25249 & 0.34164 \\ \cline{2-6} 
     & \textbf{HOMER} & 0.33531 & 0.59887 & 0.38022 & 0.39384 \\ \cline{2-6} 
     & \textbf{LP} & 0.26623 & 0.37311 & 0.43635 & 0.61094 \\ \cline{2-6} 
     & \textbf{RAkEL} & 0.55283 & 0.65243 & 0.25228 & 0.34672 \\ \hline
    \multirow{6}{*}{\textbf{SVM}} & \textbf{BR} & 0.49096 & 0.61679 & 0.28507 & 0.36258 \\ \cline{2-6} 
     & \textbf{CC} & 0.49120 & 0.61793 & 0.28621 & 0.36318 \\ \cline{2-6} 
     & \textbf{CLR} & 0.53399 & 0.64679 & 0.27885 & 0.36154 \\ \cline{2-6} 
     & \textbf{HOMER} & 0.22394 & 0.62281 & 0.54621 & 0.46839 \\ \cline{2-6} 
     & \textbf{LP} & 0.28172 & 0.36706 & 0.74164 & 0.64954 \\ \cline{2-6} 
     & \textbf{RAkEL} & 0.54850 & 0.64805 & 0.30835 & 0.35469 \\ \hline 
    \end{tabular}
    }
\end{table}


As can be seen in Table \ref{results}, the best result with corpus G1 for the micro F1 metric was established by the BR method with the LSTM algorithm. For the Hamming Loss metric, the best result was obtained by the CC problem transformation method with the RF classification algorithm. Random Forest also enabled the third, fourth and fifth best micro F1 when combined with problem transformation methods BR, CLR and RA$k$EL, respectively.


Although the RA$k$EL method obtained the third highest micro F1 for the tests performed with the RF algorithm in corpus G1, this method allowed the best results when combined with the SVM and NB algorithms. This method was tested with four different configurations for each classification algorithm used. For the RF algorithm, the best result was obtained with the creation of 14 subsets with 3 classes. For the NB algorithm, 10 subsets of 4 classes were created. The best result for the SVM algorithm was also obtained with the use of 10 subsets, but with 3 classes. The need for parameter settings to obtain the best result also occurred for the LSTM algorithm.


For the corpus G1, the best configuration of the LSTM algorithm was obtained using the first 50 words of the news represented in a 300-dimension embedding vector. The best configuration of the network has 25 neurons with a batch size of 200. The training was performed with 25 epochs with the Adam Optimizer and a learning rate of 0.01. In addition to the G1 corpus, the new corpus BFRC-PT was used. The best result was obtained with the use of the first 25 words of each news represented in an embedding matrix of 300 elements. The best configuration of the network for this corpus has 40 neurons with a batch size of 150. The training was carried out with 6 epochs and with the same function of optimization used for the corpus G1. 

The differences between the corpora used generated differences in the parameters used and in the results obtained. For the BFRC-PT corpus, the BR method with the LSTM algorithm obtained a lower micro F1 to the same method with the RF algorithm. The LSTM algorithm was also inferior to the CLR and RA$k$EL methods with the RF and SVM algorithms. Although the values obtained were lower, the $t$ test with confidence of 95\% showed that there is no statistical difference between the results obtained with the LSTM and RF algorithms. In relation to the metric hamming loss, it is possible to observe that, as for corpus G1, the best result was obtained with the CC method in conjunction with the RF algorithm. The $t$ test had showed that for this metric there is no statistical difference between the BR method with LSTM and the best result. Although the BR method with the LSTM algorithm obtained a lower result than the RA$k$EL and CLR methods with the SVM algorithm for the micro F1 metric, a higher result was recorded for the hamming loss metric. This result represents that although the LSTM obtained more correct predictions than the SVM, the distribution of the correct predictions among the classes was smaller. The best results obtained for the two corpora used have demonstrated that the strategy where several binary classifiers are created has presented more efficient than the others strategies used for the problem studied in this work.


Although the RA$k$EL method did not achieve a good result as the binary classifiers can creates, the creation of class ensembles by the RA$k$EL method enabled the third and fourth best results for the micro F1 in the BFRC-PT. As for the G1 corpus, different configurations of this method were tested for each algorithm used. For the RF and SVM algorithms, the best results were obtained with the use of 14 subsets with 3 classes. The best result for the NB algorithm was obtained with the use of 10 subsets of 4 classes. Unlike the RA$k$EL method, where different configurations were evaluated, the other methods tested were used with their default configurations or indicated settings. Among these methods are HOMER and LP, which generated the lowest results for the two corpora tested. These results demonstrate that the characteristics of corpora used make these methods less efficient.


\section{Conclusion}
\label{sec:conclusion}

In this work, we present a comparison with some problem transformation methods combined with different induction algorithms. We also present the use of a Deep Learning algorithm with a problem transformation method for the multi-label opinion mining task. The LSTM classification algorithm was used by transforming the multi-label database into several binary databases using the BR method. For the evaluation of the techniques used was introduced a new corpus of news, labeled with user reactions. The two corpora used are composed of news in Brazilian Portuguese. For the comparison of the results obtained with the proposed method and with the methods established in the literature, tests with BR, CC, CLR, HOMER, LP and RA$k$EL were performed with NB, RF and SVM algorithms and with the algorithm adaptation method ML$k$NN.


The tests performed with G1 corpus demonstrated that the combination of the LSTM algorithm with the BR method allowed the highest micro F1 among all the evaluated methods. Although this combination was most efficient, there was a difference of only 0.014pp. between the result obtained by RF using the same method. For the hamming loss metric, the best result for the two corpora was obtained with the CC method with the RF algorithm. Although the best result for the metric hamming loss was the same for both corpora, the best result obtained by the micro F1 metric for the BFRC-PT corpus was the combination between the BR method and the RF algorithm. The combination between the BR method and the LSTM algorithm enabled the sixth best result among the 20 methods tested.


The different results obtained for the different corpus used demonstrate how the characteristics of each dataset influence the choice of the method and the most appropriate algorithm. The lack of resources to classify user reactions and the high cost for the development of new resources are the main limitation for the exploration of this area, which limits the definition of a single best method or algorithm. Even though different methods of problem transformation were highlighted, it was possible to observe the high performance obtained by the strategy of creating several binary classifiers, as in BR and CC methods. Good results were also obtained with the creation of class ensemble by the RA$k$EL method. For this reason, as future work we plan to use other methods of problem transformation, especially with techniques that use a class ensemble.


\section*{Acknowledgment}
We would like to thank Siemens Ltd and NVIDIA Corporation for partially support this work.

\bibliographystyle{ACM-Reference-Format}
\bibliography{sample-sigconf.bbl}

\end{document}